# Managing contextual artificial neural networks with a service-based mediator

Greg Fish

**Abstract**

Today, a wide variety of probabilistic and expert AI systems are used to analyze real world inputs such as unstructured text, sounds, images, and statistical data. However, all these systems exist on different platforms, with different implementations, and with distinctly different, often and very specific goals in mind. This paper introduces a concept for a mediator framework for such systems and seeks to show several architectures which would support it, potential benefits in combining the signals of disparate networks for formalized, high level logic and signal processing, and its possible academic and industrial uses.

## 1. Introduction

Artificial neural networks (ANNs) have been used in artificial intelligence projects for decades and have been employed for numerous signal processing tasks. While ANNs attempt to roughly approximate the behavior of real biological systems, the sheer size and complexity of nervous systems mean that real-time full-scale simulations of entire cortexes using ANNs require supercomputing resources and are simply not practical to train to execute any exceedingly complex tasks. But even though existing ANNs have to be somewhat limited in their scope and scale, they have been successfully used to diagnose medical conditions in patients suffering from certain disorders [1], train robots how to walk using any number of limbs [2] [3], recognize human faces under controlled conditions [4], and tackle new challenges which build on tasks they've previously performed. [5]

The relatively narrow specialization of each neural network means that when a system has been trained for a particular task, multiple ANNs could be mediated by a system which can implement high level logical schemas that put the recognized signals into context and shape them into objects which can be used to make abstract decisions. According to current research into brain functions, such an approach seems to be consistent with how a biological mind seems to work. For example, neurons in the right amygdala react only to animal forms [6], the neurons in the V4 visual cortexes react to very specific features like sharp angles [7] and appear to perform feature extraction for further processing in the V5 cortex, and related neuron clusters will tend to follow Rent Scaling [8] to maximize the signal efficiency between the neurons inside them.

Furthermore, numerous experiments on insects and mice found that a highly specific brain function could be accessed simply by stimulating target clusters of neurons specially engineered to be photo-reactive to confirm that they actually perform the suspected function. [9] [10] [11] Additionally, there has even been a successful attempt to create a prosthesis which stimulates long-term memory recall as an artificial hippocampus for rats [12], and one small series of case studies has shown that a routine brain surgery which slightly damages the parietal cortex can influence the strength of the patients' religious convictions post-procedure. [13] If we see that there are networks of neurons with highly specialized purposes, the questions this fact prompts is how these discrete neural networks produce something greater than the sum of their parts.

Current theories point to these networks operating in concert and coordinated by other networks [14] which exist to make decisions using high levels of logical and formal abstraction, such as the prefrontal cortex. Extending this through further to artificial neural networks, we could imagine a framework for selecting appropriate ANNs for certain tasks or putting the functions of one or several specialized ANNs in their proper context and determining a reasonable course of action within that context. This ANN mediator would require a formalized logical layout of various concepts and its essential components to select and activate the appropriate ANN to identify an active stimulus and collect more relevant information about this stimulus based on the aforementioned formal logic.

This is what the Hivemind system is designed to do. Rather than a particular implementation of an ANN or an expert system, it's intended to be a framework for negotiating semantics and logical relationships between various concepts and a management tool to negotiate their detection and analysis. It's intended to merge and test existing and new approaches for AI components to make sense of unstructured data received from its environment. Since most signal processing systems are calculation and resource intensive, it's designed as a set of .NET WCF services able to run multiple instances simultaneously and on multiple machines. The remainder of this paper will deal with laying out the architecture and design of the Hivemind system as well as its possible applications.

**2. System Architecture Overview**

Hivemind revolves around a construct referred to as an idea. A single idea is a concept with has its attributes, performs certain actions, and has ANNs associated with it. Each action and attribute is also a concept which can be at the core of an idea object, making the relationships between them relatively simple to maintain, but leading to complex and open-ended networks of concepts that allow for high level conceptualization and a nuanced decision-making process based on numerous relevant data points supported by the framework.

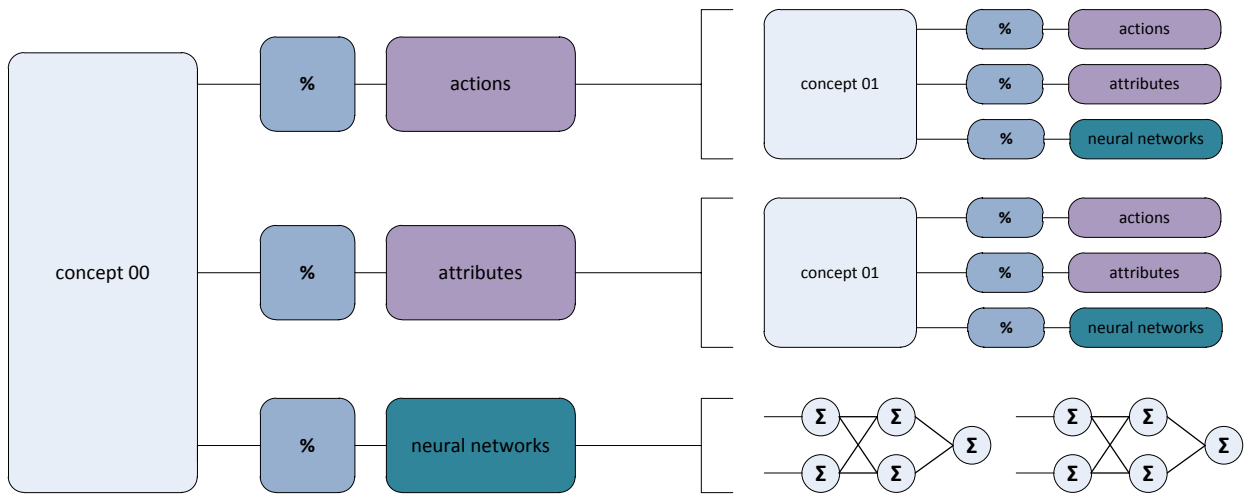

Figure 2.1 : Diagram representing the logical layout of concepts and their components in Hivemind

In addition to mapping concept to each other and specific artificial neural networks, or using an active ANN to arrive back to a core concept, Hivemind also includes a statistical strength of the relationship to add yet another layer of abstraction to an existing relationship and serve as a trigger for further system action, allowing the system employing a Hivemind service to justify its decisions and act on the input it receives from an outside actor or the environment depending on the input's context and strength. The logic of mapping and retrieving components and storing the relationships are done by the two services created by the Hivemind Web Services (.WS) component and standard stored procedures in its database (.DB) schema. The output would be available as a set of object proxies serialized via WCF using SOAP to be consumed by a wide variety of clients able to receive XML-based RPCs.

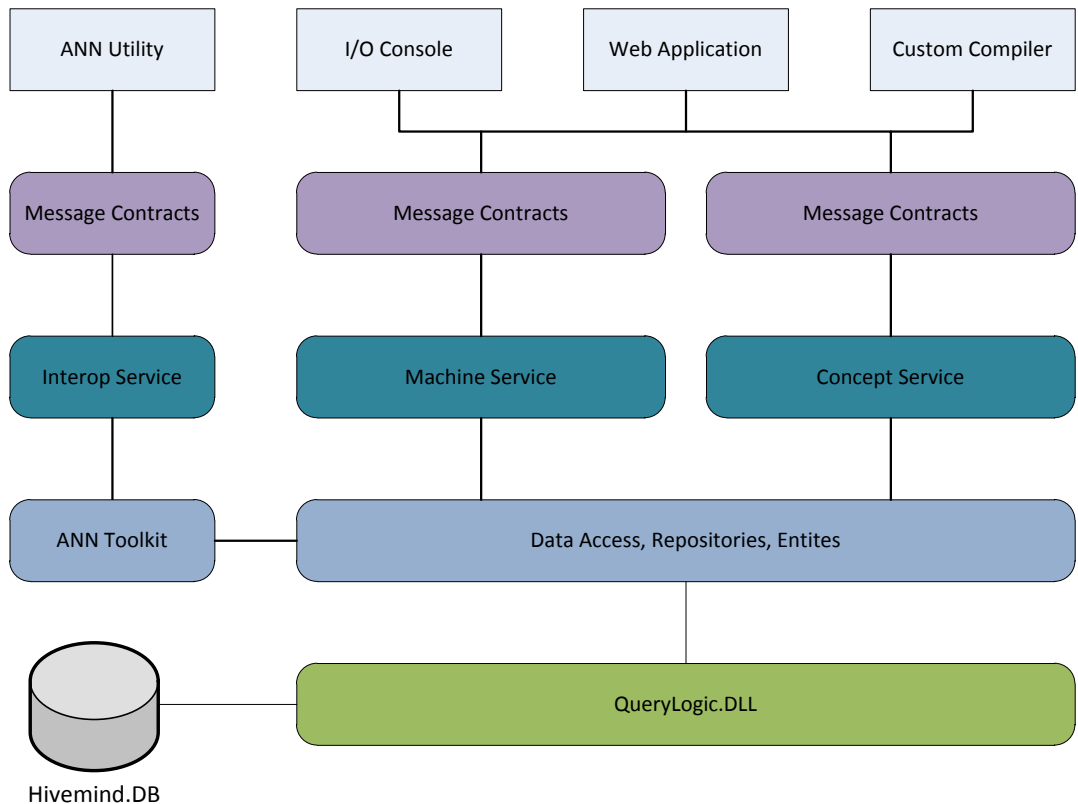

Figure 2.2 : Hivemind architectural overview schematic

Another alternative is a Hivemind UI component which presents a series of management screens using HTML and Javascript libraries to validate user input and both display and submit data using AJAX calls with JSON models for a light client footprint and better performance. This application would allow the Hivemind database to be built through Wikipedia-style crowdsourcing although this approach is not without its risks and downsides. These will be explored in the discussion section in further detail.

**2.1 Hivemind.DB Implementation**

Hivemind's database component is a relational MS SQL database using stored procedures to retrieve, save, and manipulate data. While this does have the disadvantage of marrying Hivemind to an MS SQL tool and would require modifications to existing stored procedures were this system to be redesigned to use an Oracle database, it allows Hivemind to retrieve data faster a on large scale and reduces the cost of each database hit. Using an ORM tool like Hibernate or Entity Framework would grant independence in database engines but also entail the cost of limitations to the system's scalability and produce a rather significant overhead involved in managing the objects' context. Without this context, the ORM tool would not know if the object in memory is to be stored or altered [15] and depending on the number of objects in memory at runtime, instances of the object context can grow to be a significant drain on server resources.

There are possible optimizations to speed up data retrieval using ORM tools [16] but they involve trying to modify the generated database schema to closer resemble existing optimization techniques already used in many manually modeled databases (i.e. using bridge tables, retrieving only IDs on lookups, manipulating low level ORM logic to make data manipulation behave more like a stored procedure, etc.) and a reliance on data and context caching to enable eager loading based on specific application requirements. Academically published performance tests conducted on such setups have not proven their potential for scalability since tests were ran on what could be considered a fairly typical home user setup [17] [18] and industry efforts to build massive applications using an ORM tool are still somewhat recent.

Hence, to avoid all these uncertainties and complications, Hivemind.DB does not make use of ORM tools and relies on relationships more easily maintained in a schema custom generated from an XSD document. A few NoSQL solutions were considered but this approach selected since the wide usage of Hivemind would result in a tangle of denormalized data and interfere with further work on the framework's conceptual mapping. To manage database transactions, Hivemind will use a custom database façade which provides a leaner and more efficient data retrieval than built in .NET data sets and extension methods for working with a live connection while writing less code. If the goal is to minimize connection time, the bulk retrieval mode can be used by the data access layer. When the goal is faster data retrieval, the extension methods for data readers can be used instead in another implementation of the access layer modified by users for whom speed is the outmost priority.

In the database schema itself, each mapping between related objects is only created or deleted to reduce the chance of crossed wires. Since all the concept entities already exist or are added at the time of mapping, updating would have exactly the same effect as a deletion based on a composite key consisting of a concept and its attribute or action. Because so many objects have a many to many relationship due to the modular nature of the real world concepts they represent, Hivemind.DB relies on bridge tables and employs manifesting to determine what objects, if any, the system should populate next based on what and how much data is requested by consumer applications. [19] Both the pseudo-data set and data reader implementations of QueryLogic support manifesting, however the data set implementation will handle manifesting automatically while the data reader version would require some additional code to retrieve the next set of results.

**2.2 Hivemind Services And Artificial Neural Networks**

Hivemind offers three services to its end users, one of which is the Interop Service. This service's task is to manage data for artificial neural networks in a JSON-like notation. Since the key factor in ANNs is the set of input weights which are modified by training algorithms, it's primarily the weight data that will be stored and managed to be plugged into whatever implementation is needed. The notation itself stores the values as plain ASCII text and represents each neuron as a threshold value (which is 0.5 by default) and the associated set of input weights. The activation function is then calculated at runtime, as is usually done. When each neuron is decoded, it can be linked to a particular sensor or motor via a consuming application and allow users to train a machine.

In this notation, the neural network values would be saved in a field of the ANN relation of the database. The goal is to make it as easy as possible to work with ANNs regardless of how they were implemented and free highly accurate implementations from their source code. A library for unpacking and running fully trained ANNs as well as training any new ones, then packaging them into Hivemind's notation could easily be referenced in the consuming application to reduce the amount of data sent over the wire, or used by the Interop service to encode and decode ANN data before sending it to, or receiving it from, the consumer application. Which method is used will depend on how the framework will be used and with what resources but current implementation can support both.

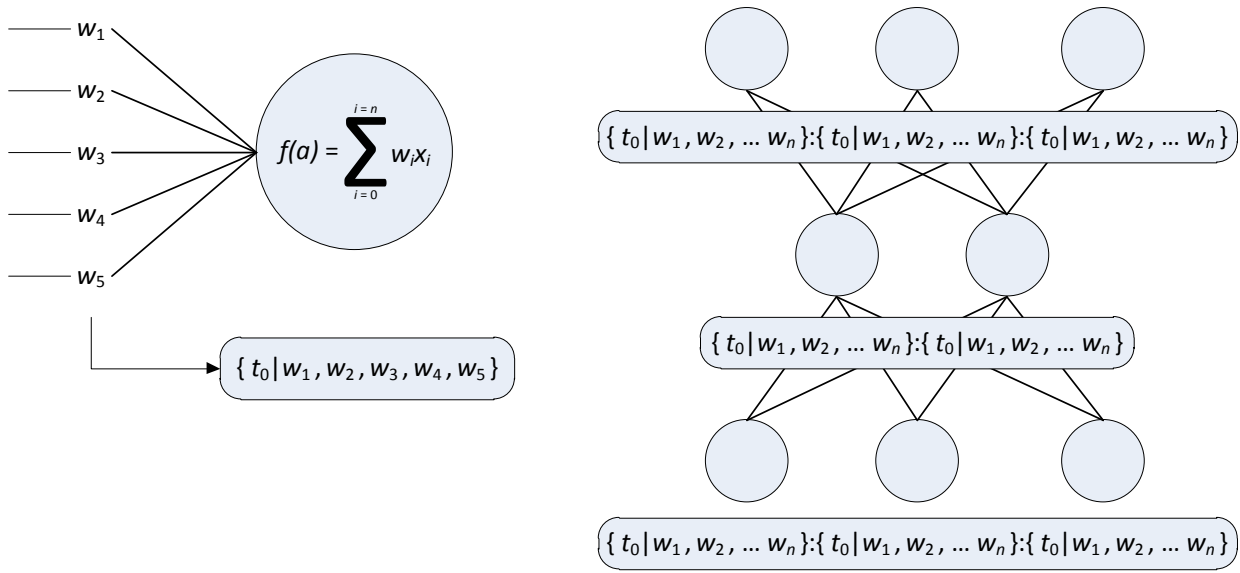

Figure 2.3 : Conceptual diagram of the JSON-like ASCII-based notation for artificial neural networks. Each layer consists of *n* neurons represented by a set of weights with a specified threshold value for the neuron to fire.

**2.3 Hivemind.WS Implementation**

Hivemind.WS follows fairly straightforward SOA architecture and performs little business logic by itself since its goal is to efficiently retrieve and organize data as specified by the Hivemind.DB schema to give the framework's users the resources to manipulate the relevant ANNs and logical formalisms. What it will provide are pieces to be put together through experimentation or prior knowledge since each unit (i.e. an idea or a concept, or a machine, or motors) are treated as modular, hence the many to many relationships being mediated during data storage and retrieval. Hivemind.WS is built to support a robot with a practically unlimited number of motors and sensors each with a practically unlimited number of commands if we were to temporarily ignore data type value limitations and database constraints and focus only on a practical usage scenario. This gives users flexibility to use the framework with any machine of their choosing rather than being locked in to a particular platform. Most of the customizations discussed in this paper could happen by simply exchanging or editing the different parts of the DLL stack above the data access, entity, and data contract layers.

To control the amount of information to be managed by a client, Hivemind.WS' message contracts are built to support manifesting. For an example, if a consumer only needs to see the basic information about the various machines tracked by the system, it would format its request not to include a call for additional motor objects and all of their command and argument objects. Alternatively, when the client needs the additional information, it can specify a collection of string objects which will be used to build queries which will retrieve the data for additional objects, build them for retrieval, and send them with the response. Hivemind.WS also implements a number of features to help speed up data retrieval and manipulation which can be further fine tuned for specific systems on which it will run utilizing the ANN and concept data in order to properly queue the best artificial neural networks to perform detection and response, setting up a logical loop for scripted instructions conveyed to a consumer that references the Hivemind.WS client library responsible for communicating with the services and creating the action and concept maps for the supported concepts, and making them available for the adapter for client machines.

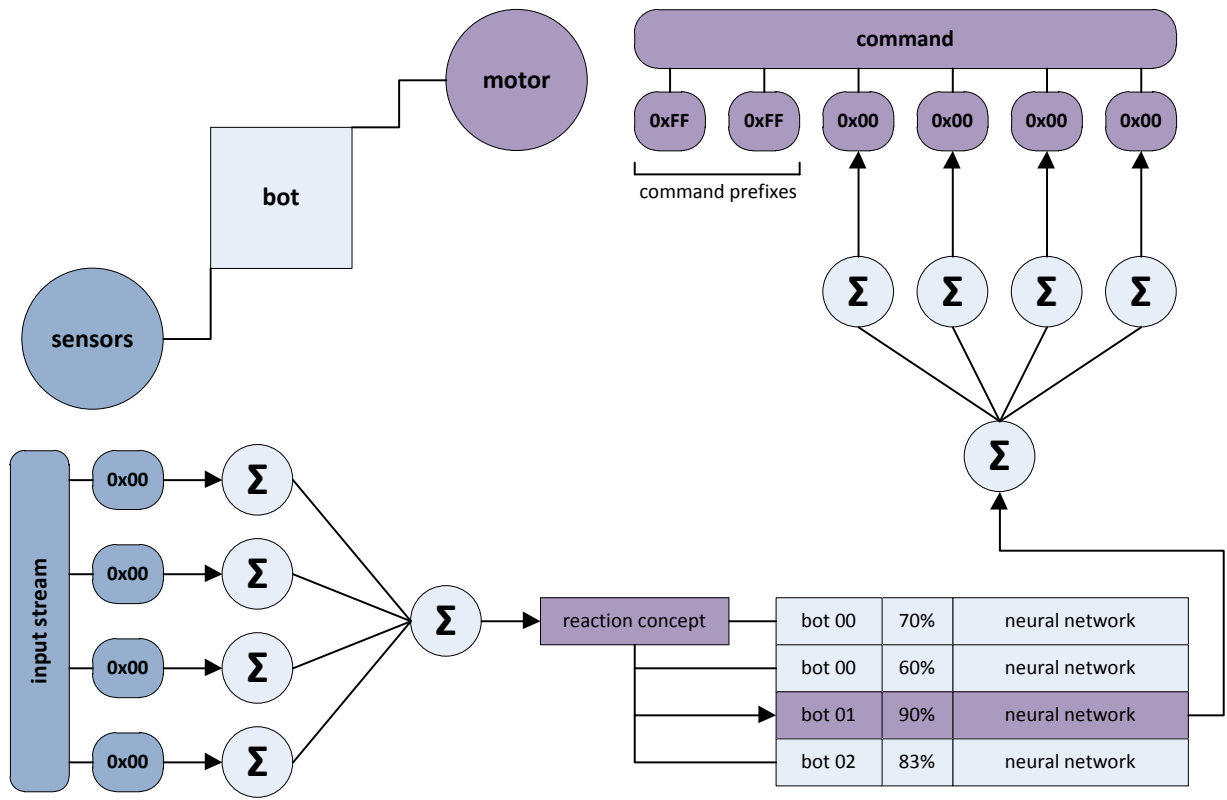

Figure 2.4 : Logical schematic of desired detection-response cycle in a client application. Groups of visual, audio, pressure, and other input sensors feed data to artificial neurons which determine if an identification has been made, then feeds the result into a proper ANN which will create the byte code arguments for a robot's motors.

Since each robotics platform is unique and uses its own communication protocols, Hivemind.WS would not be able to directly control a particular machine, however the modular nature of the technology stack on which it's built allows it to enough information for an adapter intended to interface with a particular robot via the Machine Service along with the ANNs required to make it act accordingly through the Concept Service. The adapter would essentially match up the inputs from sensors with an ANN's input neurons which will then allow the ANN's output to be fed to a response-generating ANN the output of which would match the proper argument structure for the robot's motors. This method would allow for a more varied and subtle response to either an affirmative detection of sought objects, or being unable to find them as well as identifying sought objects or input patterns in general. Should the main concept be complex or difficult to identify for whatever reason, the machine could load the best ANNs to identify its attributes (static qualities) or actions (motion patterns) to hone in on the sought object. The object's relationship to other concepts and the relationship's strength can then be used to further advance a chain of actions and responses for individual machines and even entire robot swarms since Hivemind is focused not on a particular robot's abilities but rather selecting the right machines for the tasks specified.

**3. Swarm Control And Robot Autonomy With Hivemind.WS**

Teaching machines how to navigate the real world and make decisions based on high order logic often tends to take the shape of creating artificial neural networks (ANNs) which learn tasks by trial and error, or rigid expert systems which interact with their environments in preset ways. ANNs can produce wide ranges of behavior and settle on adequate solutions to problems a machine encounters in a given environment and allow the machines to retrain themselves if a new obstacle arises. This methodology can even give rise to complex behaviors exhibited by animals and insects in nature [20] and produce two radically different approaches within the same group of robots implementing the same ANN structures with different inputs [21] [22]. But this complexity and originality on the part of the machines can come with a degree of unpredictability, especially when a new stimulus is introduced into the environment. There's always a chance that a probabilistic system will take a non-optimal course of action or even that it will become stuck as a result of over-fitting and fail to adapt.

This is why in situations where machines have to follow a set of steps with no room for innovation to perform a sensitive or critical task (be it on a car assembly line or disarming a bomb), they are either programmed to follow an exacting routine, or to cede control to humans. At the same time, however, a purely routinized approach means that changes to the machine's behavior have to be implemented in very explicit code and subjected to many hours of testing and modification. Likewise, the machine will have a very limited, if any, understanding of its environment and unable to implement any benefits of true autonomy such as new ways of accomplishing a task or dealing with minor obstacles that may phase a more autonomous robot. To bridge the major differences between the ANN and an expert system approach, there is the Structure framework. Built on the idea of leveraging concepts in carefully mapped recursive relationships between each other, external stimuli, and ANNs.

At the core of a client consuming Hivemind.WS services will be action and concept hash maps which host the implementations of various behaviors in ANNs associated with specified keywords or supporting the data used for triggering and managing complex behaviors. Those keywords are concepts and the implementation to which they are mapped supply the machine with a proper set of values for an ANN to load in order to detect the feature or the action in question, or a set of steps to take when some decision regarding the sensory inputs has been made, and an additional ANN in response to the detection as per user specifications. As new keywords come in from users or scripts, or related concepts surface, Hivemind can asynchronously supply robots with the relevant implementation until a goal has been reached.

This structure also allows users to run experiments testing how well particular ANNs or expert settings would perform in the real world by monitoring the accuracy of their detections, complete with a log of their efforts. Any underperforming or overfitted ANNs can be flushed out and replaced with updated and more accurate versions via a download from Hivemind.WS. When dealing with multiple machines of different types, the client can deploy the latest and most accurate ANNs and settings in local hubs that would be able to detect far more information about their environment, and synchronized to work in concert simply by leveraging the relationships between concepts and machines created by Hivemind.WS. Additionally, the logical mappings between robots, concepts, and ANNs in Hivemind.WS allows users to grade the effectiveness of any new ANNs and use concept names and descriptions to define not only the right machines and ANNs for a set of tasks, but also the most effective machine configurations automatically next time the tasks are invoked since the selections are governed by the mappings' efficacy.

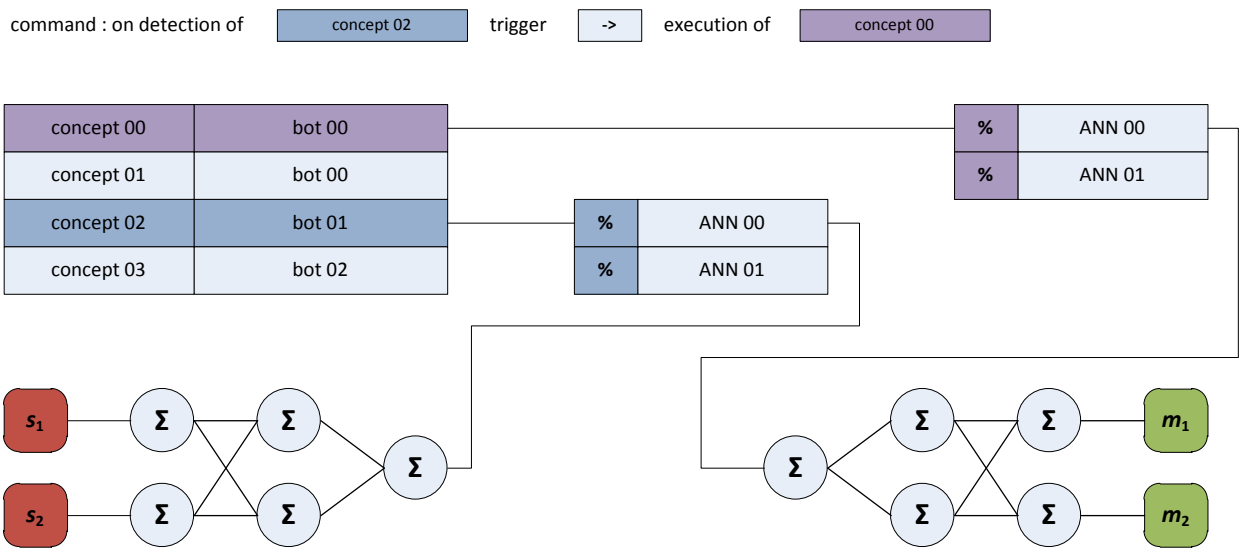

Figure 3.1 : Logical diagram of how user specified concepts are matched with proper machines and ANNs

Each action can be defined by a machine specific ANN since it is relatively straightforward to implement an ANN responsible for robot motions and fine-tuning them with each sensor and joint functioning as a neuron. [23] Again, the goal is to provide a framework for accomplishing a variety of tasks and allowing suitable machines to better work in complex environment rather than create a particular methodology for a very narrow set of tasks. Using Hivemind.WS allows to not only implement unique out-of-the-box approaches to tackling machine behavior and cognition and test their detection mechanisms for better efficiency and accuracy, but to experiment with multiple methodologies, technologies, and approaches towards navigating the real world in a semi-autonomous manner with the deployed robots acting as a single super-organism striving towards the same goals and controlled by a single artificial brain.

**4. Future advanced swarm control with Hivemind.WS**

One of the most exciting and recently active areas in robotics research is swarm control, defined here as the ability to command an entire fleet of machines to achieve a common goal. There is a variety of approaches on the mechanics of a robotic swarm ranging from disparate but cooperating robots [24] [25], to very general algorithms assigning a set of positions for machines within a network and maintaining their formations during a mission. [26] [27] While both approaches are valid depending on the context and situations in which they are deployed, existing literature makes it somewhat difficult to imagine scaling up current small-scale experiments and simulations into large, real-world operations. A major part of the difficulty is allowing machines to a navigate complex, real-world environment and the fact that numerous projects in advancing robotic autonomy are focused on relatively narrow implementations, such as navigating in rooms or making their way through clutter or generic approaches to inter-robot communication. This complexity of environment is one of the issues Hivemind.WS tackles directly through its concept-machine-ANN mappings.

Again, all of these strategies are valid when dealing with relatively small or well defined spaces where basic environmental variables can be well defined and controlled since such space can be navigated using sets of SLAM algorithms which consider various robot configurations [28] [29] and learning alongside human interactions with the machine trying to navigate its environment. [30] However, a task such as moving through a city or covering a large amount of terrain from the air without any human intervention cannot be implemented using a completely autonomous, piecemeal exploration of the machine's surroundings. A standardized and precise coordinate system such as latitude, longitude, and altitude coordinates would allow machines focus only on what is and isn't relevant to getting from point A to point B and implement their arrays of detection strategies in limited environments on a step-by-step basis. Using action and concept mappings in the Hivemind framework, machine given geo-coordinates as a goal can either follow a user-created script to manage its navigation process using an API, or load a default set of actions to reach its user-specified target destination.

Rather than pre-calculating the best set of steps to get where it's going and then distributing these paths and settings to similar machines within its network as some approaches suggest [31], Hivemind would allow machines under its controls to processes the path, the steps needed to be taken to follow its chosen path, and the available alternatives long the way asynchronously. The implementations can be encoded using the aforementioned API and the actions themselves dispatched through the action and function maps in the business logic layers of a Hivemind expansion library. In addition to providing this flexibility, the system's planned ability to know where all units it will control are located and display them on a map opens the door to a human who can give high level commands to a machine on a significant scale while a change to a robot in one city will be known by machines located in a nearby suburb or across the continent, and taken into account if its location plays a role in the requested task. The overall goal is to effectively network together machines capable of asynchronous context-aware detection and operation, leverage their existing successful approaches to signal processing, and give humans control of their positions and actions on the fly rather than preprogramming them for every single motor action in advance.

**5. Discussion**

Enabled with a logical map of a particular concept, Hivemind could be deployed in a variety of ways to help machines explore and make conclusions about their surroundings or navigate unstructured text intended for other humans. For example, if machines detect that they are within four walls and there is a roof overhead, Hivemind's Concept Service could tell them that they are within a building and if they'd like to exit, they need to find a door. It can also specify that a door has a knob and by detecting the shape of a knob with their sensors, they would know where the exit is and either line up to leave when the door is opened, or alert a human where the exit might be in an emergency. Essentially, it would give machine the ability to make conclusions about the context of what it was detecting without the requirement that it detects an entire object. Just detecting a component or an action would be enough to produce a list of possibilities the machine can then explore and then plug into a logic circuit using the Gaussian values grading the strength of the relationships between the concepts to suggest new objects to detect for difficult environments, or evaluate the probability that the identification was correct.

These abilities could also be the basis of higher level logic for autonomous machines and allow them to make more decisions about the environment around them and how to negotiate it. This is meant to be the primary area of work for the Hivemind framework. Besides a natural or an artificial environment, there may be another situation in which computers would be able to use Hivemind to find their way around. Search engine spiders crawl the web to index the contents of billions of pages of primarily unstructured content, relying on a wide multitude of factors to evaluate the accuracy of their results. A formal knowledge of a particular concept and what it entails would help them navigate an unstructured black of text by making sure that the words on the page appear in proper context and with an adequate frequency to be a valuable piece of content relevant to the search rather than a spam page or a short, barely relevant result so often found during highly specialized and technical searches.

A further benefit of this method of parsing text is that it wouldn't really require the engine to have any prior knowledge about the popularity of the URL and who links to it. Many web searches tend to lead to highly popular sites in a specific niche which may not give the most relevant information for the searcher and a more obscure site may offer more relevant and useful information. This algorithm based on the Hivemind framework could be built on comparing relevant, irrelevant, and barely relevant sample to fine-tune the logic circuit of the search spider and properly index the page. Though this scenario brings us to a challenge in implementing Hivemind. Its database has to be properly populated and its concepts would have to be accurately mapped. There are several ways to do this, the most reliable of which are either SQL insert scripts with vetted data, or an I/O console application which calls service methods with data from a text or custom dictionary with the required relationships. Another option would involve using a website like Wikipedia and crawling through its relationship trees to build up the necessary maps, but this will entail significant intervention to make sure the mapping logic is correctly implemented.

Finally, another way to populate Hivemind's conceptual formalism is through the crowdsourcing efforts that were mentioned previously, however, the crowd in question will really have to understand the tasks at hand and be independent enough to peer-review each others' work. In crowdsourcing tasks, poorly managed projects tend to produce a high rate of failure in even simple tasks [32] and transparency in crowdsourcing projects could make the participants prone to groupthink [33], leading to exacerbation of even the smallest errors. However, a massive crowdsourcing approach would allow for far more information to be entered into Hivemind and curated to ensure proper relationships. Which approach is best and how its results should be best applied to real-life problems could be topics to explore in further research.